# Unification of Fusion Theories (UFT)

Florentin Smarandache
Department of Mathematics and Sciences
University of New Mexico
200 College Road
Gallup, NM 87301, USA
smarand@unm.edu

**Abstract:** Since no fusion theory neither rule fully satisfy all needed applications, the author proposes a Unification of Fusion Theories and a combination of fusion rules in solving problems/applications. For each particular application, one selects the most appropriate model, rule(s), and algorithm of implementation.
We are working in the unification of the fusion theories and rules, which looks like a cooking recipe, better we'd say like a logical chart for a computer programmer, but we don't see another method to comprise/unify all things.
The unification scenario presented herein, which is now in an incipient form, should periodically be updated incorporating new discoveries from the fusion and engineering research.

## 1. Introduction.

Each theory works well for some applications, but not for all.
We extend the power and hyper-power sets from previous theories to a Boolean algebra obtained by the closure of the frame of discernment under union, intersection, and complement of sets (for non-exclusive elements one considers as complement the fuzzy or neutrosophic complement). All bbas and rules are redefined on this Boolean algebra.
A similar generalization has been previously used by Guan-Bell (1993) for the Dempster-Shafer rule using propositions in sequential logic, herein we reconsider the Boolean algebra for all fusion rules and theories but using sets instead of propositions, because generally it is harder to work in sequential logic with summations and inclusions than in the set theory.

## 2. Fusion Space.

For $n \geq 2$ let $\Theta = \{\theta_1, \theta_2, \ldots, \theta_n\}$ be the frame of discernment of the fusion problem/application under consideration. Then $(\Theta, \cup, \cap, \mathcal{C})$, $\Theta$ closed under these three operations: union, intersection, and complementation of sets respectively, forms a Boolean algebra. With respect to the partial ordering relation, the inclusion $\subseteq$, the minimum element is the empty set $\phi$, and the maximal element is the total ignorance $I = \bigcup_{i=1}^{n} \theta_i$.

Similarly one can define: $(\Theta, \cup, \cap, \setminus)$ for sets, $\Theta$ closed with respect to each of these operations: union, intersection, and difference of sets respectively.
$(\Theta, \cup, \cap, \mathcal{C})$ and $(\Theta, \cup, \cap, \setminus)$ generate the same super-power set $S^\Theta$ closed under $\cup, \cap, \mathcal{C}$, and $\setminus$ because for any $A, B \in S^\Theta$ one has $\mathcal{C}A = I \setminus A$ and reciprocally $A \setminus B = A \cap \mathcal{C}B$.

If one considers propositions, then $(\Theta, \vee, \wedge, \neg)$ forms a Lindenbaum algebra in sequential logic, which is isomorphic with the above $(\Theta, \cup, \cap, \mathcal{C})$ Boolean algebra.

By choosing the frame of discernment $\Theta$ closed under $\cup$ only one gets DST, Yager's, TBM, DP theories. Then making $\Theta$ closed under both $\cup, \cap$ one gets DSm theory. While, extending $\Theta$ for closure under $\cup, \cap$, and $\mathcal{C}$ one also includes the complement of set (or negation of proposition if working in sequential logic); in the case of non-exclusive elements in the frame of discernment one considers a fuzzy or neutrosophic complement. Therefore the super-power set $(\Theta, \cup, \cap, \mathcal{C})$ includes all the previous fusion theories.

The power set $2^\Theta$, used in DST, Yager's, TBM, DP, which is the set of all subsets of $\Theta$, is also a Boolean algebra, closed under $\cup, \cap$, and $\mathcal{C}$, but does not contain intersections of elements from $\Theta$.
The Dedekind distributive lattice $D^\Theta$, used in DSmT, is closed under $\cup, \cap$, and if negations/complements arise they are directly introduced in the frame of discernment, say $\Theta'$, which is then closed under $\cup, \cap$. Unlike others, DSmT allows intersections, generalizing the previous theories.
The Unifying Theory contains intersections and complements as well.

Let's consider a frame of discernment $\Theta$ with exclusive or non-exclusive hypotheses, exhaustive or non-exhaustive, closed or open world (all possible cases).

We need to make the remark that in case when these $n \geq 2$ elementary hypotheses $\theta_1, \theta_2, \ldots, \theta_n$ are *exhaustive and exclusive* one gets the Dempster-Shafer Theory, Yager's, Dubois-Prade Theory, Dezert-Smarandache Theory, while for the case when the hypotheses are *non-exclusive* one gets Dezert-Smarandache Theory, but for *non-exhaustivity* one gets TBM. An exhaustive frame of discernment is called *close world*, and a non-exhaustive frame of discernment is called *open world* (meaning that new hypotheses might exist in the frame of discernment that we are not aware of). $\Theta$ may be finite or infinite.

Let $m_j: S^\Theta \rightarrow [0, 1]$, $1 \leq j \leq s$, be $s \geq 2$ basic belief assignments,
(when bbas are working with crisp numbers),
or with subunitary subsets, $m_j: S^\Theta \rightarrow \mathcal{P}([0, 1])$, where $\mathcal{P}([0, 1])$ is the set of all subsets of the interval $[0,1]$ (when dealing with very imprecise information).

Normally the sum of crisp masses of a bba, $m(.)$, is 1, i.e. $\sum_{X \in S^\Theta} m(X) = 1$.

## 3. Incomplete and Paraconsistent Information.

For incomplete information the sum of a bba components can be less than 1 (not enough information known), while in paraconsistent information the sum can exceed 1 (overlapping contradictory information).
The masses can be normalized (i.e. getting the sum of their components =1), or not (sum of components < 1 in incomplete information; or > 1 in paraconsistent information).

For a bba valued on subunitary subsets one can consider, as normalization of m(.),
either $\sum_{X \in S^{\wedge T}} \sup\{m(X)\} = 1$,

or that there exist crisp numbers $x \in X$ for each $X \in S^\Theta$ such that $\sum_{\substack{X \in S^{\wedge T} \\ x \in X}} m(x) = 1$.

Similarly, for a bba m(.) valued on subunitary subsets dealing with paraconsistent and incomplete information respectively:
a) for incomplete information, one has $\sum_{X \in S^{\wedge T}} \sup\{m(X)\} < 1$,

b) while for paraconsistent information one has $\sum_{X \in S^{\wedge T}} \sup\{m(X)\} > 1$ and there do not exist crisp numbers $x \in X$ for each $X \in S^\Theta$ such that $\sum_{\substack{X \in S^{\wedge T} \\ x \in X}} m(x) = 1$.

## 4. Specificity Chains.

We use the min principle and the precocious/prudent way of computing and transferring the conflicting mass.

Normally by transferring the conflicting mass and by normalization we diminish the specificity.
If A∩B is empty, then the mass is moved to a less specific element A (also to B), but if we have a pessimistic view on A and B we move the mass m(A∩B) to A∪B (entropy increases, imprecision increases), and even more if we are very pessimistic about A and B: we move the conflicting mass to the total ignorance in a closed world, or to the empty set in an open world.

Specificity Chains:
a) From specific to less and less specific (in a closed world):
(A∩B) ⊂ A ⊂ (A∪B) ⊂ I or (A∩B) ⊂ B ⊂ (A∪B) ⊂ I.
Also from specific to unknown (in an open world):
A∩B → ϕ.
b) And similarly for intersections of more elements: A∩B∩C, etc.
A∩B∩C ⊂ A∩B ⊂ A ⊂ (A∪B) ⊂ (A∪B∪C) ⊂ I
or (A∩B∩C) ⊂ (B∩C) ⊂ B ⊂ (A∪B) ⊂ (A∪B∪C) ⊂ I, etc. in a closed world.
Or A∩B∩C → ϕ in an open world.
c) Also in a closed world:
A∩(B∪C) ⊂ B∪C ⊂ (B∪C) ⊂ (A∪B∪C) ⊂ I or A∩(B∪C) ⊂ A ⊂ (A∪B) ⊂ (A∪B∪C) ⊂ I.

Or A∩(B∪C) → ϕ in an open world.

## 5. Static and Dynamic Fusion.

According to Wu Li we have the following classification and definitions:
*Static fusion* means to combine all belief functions simultaneously.
*Dynamic fusion* means that the belief functions become available one after another sequentially, and the current belief function is updated by combining itself with a newly available belief function.

## 6. Scenario of Unification of Fusion Theories.

Since everything depends on the application/problem to solve, this scenario looks like a logical chart designed by the programmer in order to write and implement a computer program, or like a cooking recipe.

Here it is the scenario attempting for a unification and reconciliation of the fusion theories and rules:

1) If all sources of information are reliable, then apply the conjunctive rule, which means consensus between them (or their common part):
2) If some sources are reliable and others are not, but we don't know which ones are unreliable, apply the disjunctive rule as a cautious method (and no transfer or normalization is needed).
3) If only one source of information is reliable, but we don't know which one, then use the exclusive disjunctive rule based on the fact that $X_1 \veebar X_2 \veebar \ldots \veebar X_n$ means either $X_1$ is reliable, or $X_2$, or and so on or $X_n$, but not two or more in the same time.
4) If a mixture of the previous three cases, in any possible way, use the mixed conjunctive-disjunctive rule.
As an example, suppose we have four sources of information and we know that: either the first two are telling the truth or the third, or the fourth is telling the truth.
The mixed formula becomes:
$m_{\cap\cup}(\phi) = 0$, and $\forall A \in S^\Theta \setminus \phi$, one has $m_{\cap\cup}(A) = \sum_{\substack{X_1, X_2, X_3, X_4 \in S^{\wedge\Theta} \\ ((X_1 \cap X_2) \cup X_3) e \cup X_4 = A}} m_1(X_1) m_2(X_2) m_3(X_3) m_4(X_4)$.

5) If we know the sources which are unreliable, we discount them. But if all sources are fully unreliable (100%), then the fusion result becomes vacuum bba (i.e. $m(\Theta) = 1$, and the problem is indeterminate. We need to get new sources which are reliable or at least they are not fully unreliable.
6) If all sources are reliable, or the unreliable sources have been discounted (in the <u>default case</u>), then use the DSm classic rule (which is commutative, associative, Markovian) on Boolean algebra ($\Theta, \cup, \cap, \mathcal{C}$), no matter what contradictions (or model) the problem has. I emphasize that the super-power set $S^\Theta$ generated by this Boolean algebra contains singletons, unions, intersections, and complements of sets.
7) If the sources are considered from a statistical point of view, use *Murphy's average rule* (and no transfer or normalization is needed).
8) In the case the model is not known (the <u>default case</u>), it is prudent/cautious to use the free model (i.e. all intersections between the elements of the frame of discernment are non-empty) and DSm classic rule on $S^\Theta$, and later if the model is found out (i.e. the constraints of empty intersections become known), one can adjust the conflicting mass at any time/moment using the DSm hybrid rule.
9) Now suppose the model becomes known [i.e. we find out about the contradictions (= empty intersections) or consensus (= non-empty intersections) of the problem/application]. Then :
    9.1)    If an intersection $A \cap B$ is not empty, we keep the mass $m(A \cap B)$ on $A \cap B$, which means consensus (common part) between the two hypotheses A and B (i.e. both hypotheses A and B are right) [here one gets *DSmT*].
    9.2)    If the intersection $A \cap B = \emptyset$ is empty, meaning contradiction, we do the following :
       9.2.1) if one knows that between these two hypotheses A and B one is right and the other is false, but we don't know which one, then one transfers the mass $m(A \cap B)$ to $m(A \cup B)$, since $A \cup B$ means at least one is right [here one gets *Yager's* if n=2, or *Dubois-Prade*, or *DSmT*];
       9.2.2) if one knows that between these two hypotheses A and B one is right and the other is false, and we know which one is right, say hypothesis A is right and B is false, then one transfers the whole mass $m(A \cap B)$ to hypothesis A (nothing is transferred to B);
       9.2.3) if we don't know much about them, but one has an optimistic view on hypotheses A and B, then one transfers the conflicting mass $m(A \cap B)$ to A and B (the nearest specific sets in the Specificity Chains) [using *Dempster's*, *PCR2-5*]
       9.2.4) if we don't know much about them, but one has a pessimistic view on hypotheses A and B, then one transfers the conflicting mass $m(A \cap B)$ to $A \cup B$ (the more pessimistic the further one gets in the Specificity Chains: $(A \cap B) \subset A \subset (A \cup B) \subset I$); this is also the <u>default case</u> [using *DP's*, *DSm hybrid rule*, *Yager's*];
if one has a very pessimistic view on hypotheses A and B then one transfers the conflicting mass $m(A \cap B)$ to the total ignorance in a closed world [*Yager's*, *DSmT*], or to the empty set in an open world [*TBM*];
       9.2.5.1) if one considers that no hypothesis between A and B is right, then one transfers the mass $m(A \cap B)$ to other non-empty sets (in the case more hypotheses do exist in the frame of discernment) - different from A, B, $A \cup B$ - for the reason that: if A and B are not right then there is a bigger chance that other hypotheses in the frame of discernment have a higher subjective probability to occur; we do this transfer in a **closed world** [DSm hybrid rule]; but, if it is an **open world**, we can transfer the mass $m(A \cap B)$ to the empty set leaving room for new possible hypotheses [here one gets *TBM*];
       9.2.5.2) if one considers that none of the hypotheses A, B is right and no other hypothesis exists in the frame of discernment (i.e. n = 2 is the size of the frame of discernment), then one considers the **open world** and one transfers the mass to the empty set [here *DSmT* and *TBM* converge to each other].

Of course, this procedure is extended for any intersections of two or more sets: $A \cap B \cap C$, etc. and even for mixed sets: $A \cap (B \cup C)$, etc.

If it is a dynamic fusion in a real time and associativity and/or Markovian process are needed, use an algorithm which transforms a rule (which is based on the conjunctive rule and the transfer of the conflicting mass) into an associative and Markovian rule by storing the previous result of the conjunctive rule and, depending of the rule, other data. Such rules are called quasi-associative and quasi-Markovian.

Some applications require the necessity of **decaying the old sources** because their information is considered to be worn out.

If some bba is not normalized (i.e. the sum of its components is < 1 as in incomplete information, or > 1 as in paraconsistent information) we can easily divide each component by the sum of the components and normalize it. But also it is possible to fusion incomplete and paraconsistent masses, and then normalize them after fusion. Or leave them unnormalized since they are incomplete or paraconsistent.

PCR5 does the most mathematically exact (in the fusion literature) redistribution of the conflicting mass to the elements involved in the conflict, redistribution which exactly follows the tracks of the conjunctive rule.

### 7. Examples:

#### 7.1. Bayesian Example:
Let $\Theta = \{A, B, C, D, E\}$ be the frame of discernment.

|  | A | B | C | D | E | A∩B | A∩C | A∩D | A∩E | B∩C | B∩D |
|---|---|---|---|---|---|---|---|---|---|---|---|
|  |  |  |  |  |  | ≠ φ | = φ | = φ | = φ | Not known if = or ≠ φ | = φ |
|  |  |  |  |  |  | Consensus between A and B | Contradiction between A and C, but optimistic in both of them | One right, one wrong, but don't know which one | A is right, B is wrong | Don't know the exact model | Unknown any relation between B and D. |
| $m_1$ | 0.2 | 0 | 0.3 | 0.4 | 0.1 |  |  |  |  |  |  |
| $m_2$ | 0.5 | 0.2 | 0.1 | 0 | 0.2 |  |  |  |  |  |  |
| $m_{12}$ | 0.10 | 0 | 0.03 | 0 | 0.02 | 0.04 | 0.17 | 0.20 | 0.09 | 0.06 | 0.08 |
|  |  |  |  |  |  | ↓ | ↓ | ↓ | ↓ | ↓ | ↓ |
|  |  |  |  |  |  | A∩B | A, C | A∪B | A | B∩C We keep the mass 0.06 on B∩C till we find out more information on the model. | B∪D |
| $m_r$ |  |  |  |  |  | 0.04 | 0.107, 0.063 | 0.20 | 0.09 | 0.06 | 0.08 |
| **$m_{UFT}$** | **0.324** | **0.040** | **0.119** | **0** | **0.027** | **0.04** | **0** | **0** | **0** | **0.06** | **0** |
| $m_{lower}$ (closed world) | 0.10 | 0 | 0.03 | 0 | 0.02 |  |  |  |  |  |  |

| | | | | | | | | | | |
|---|---|---|---|---|---|---|---|---|---|---|
| $m_{lower}$ (open world) | 0.10 | 0 | 0.03 | 0 | 0.02 | | | | | |
| $m_{middle}$ (default) | 0.10 | 0 | 0.03 | 0 | 0.02 | | | | | |
| $m_{upper}$ | 0.400 | 0.084 | 0.178 | 0.227 | 0.111 | | | | | |

Table 1. Bayesian Example using the Unified Fusion Theories rule regarding a mixed redistribution of partial conflicting masses (Part 1).

| | B∩E | C∩D | C∩E | D∩E | A∪B | A∪C | A∪D | A∪E | B∪C |
|---|---|---|---|---|---|---|---|---|---|
| | ≠ ϕ | = ϕ | = ϕ | = ϕ | | | | | |
| | The intersection is not empty, but neither B∩E nor B∪E interest us | Pessimistic in both C and D | Very pessimistic in both C and E | Both D and E are wrong | | | | | |
| $m_1$ | | | | | | | | | |
| $m_2$ | | | | | | | | | |
| $m_{12}$ | 0.02 | 0.04 | 0.07 | 0.08 | | | | | |
| | ↓ B, E | ↓ C∪D | ↓ A∪B∪C∪D∪E | ↓ A,B,C | | | | | |
| $m_r$ | 0.013, 0.007 | 0.04 | 0.07 | 0.027, 0.027, 0.027 | | | | | |
| **$m_{UFT}$** | **0** | **0** | **0** | **0** | **0** | **0** | **0.20** | **0** | **0** |
| $m_{lower}$ (closed world) | | | | | | | | | |
| $m_{lower}$ (open world) | | | | | | | | | |
| $m_{middle}$ (default) | | | | | | 0.04 | 0.17 | 0.20 | 0.09 | 0.06 |
| $m_{upper}$ | | | | | | | | | |

Table 1. Bayesian Example using the Unified Fusion Theories rule regarding a mixed redistribution of partial conflicting masses (Part 2).

| | B∪D | B∪E | C∪D | C∪E | D∪E | A∪B∪C∪D∪E | ϕ |
|---|---|---|---|---|---|---|---|
| | | | | | | | |
| | | | | | | | |
| $m_1$ | | | | | | | |
| $m_2$ | | | | | | | |
| $m_{12}$ | | | | | | | |
| | | | | | | | |
| $m_r$ | | | | | | | |
| **$m_{UFT}$** | **0.08** | **0** | **0.04** | **0** | **0** | **0.07** | **0** |
| $m_{lower}$ | | | | | | 0.85 | |

| (closed world) $m_{lower}$ | | | | | | 0.85 |
|---|---|---|---|---|---|---|
| (open world) $m_{middle}$ (default) | 0.08 | 0.02 | 0.04 | 0.07 | 0.08 | |
| $m_{upper}$ | | | | | | |

Table 1. Bayesian Example using the Unified Fusion Theories rule regarding a mixed redistribution of partial conflicting masses (Part 3).

We keep the mass $m_{12}(B \cap C) = 0.06$ on $B \cap C$ (eleventh column in Table 1, part 1) although we don't know if the intersection $B \cap C$ is empty or not (this is considered the default model), since in the case when it is empty one considers an open world because $m_{12}(\phi)=0.06$ meaning that there might be new possible hypotheses in the frame of discernment, but if $B \cap C \neq \phi$ one considers a consensus between B and C.
Later, when finding out more information about the relation between B and C, one can transfer the mass 0.06 to $B \cup C$, or to the total ignorance I, or split it between the elements B, C, or even keep it on $B \cap C$.

$m_{12}(A \cap C)=0.17$ is redistributed to A and C using the PCR5:
a1/0.2 = c1/0.1 = 0.02(0.2+0.1),
whence a1 = 0.2(0.02/0.3) = 0.013,
c1 = 0.1(0.02/0.3) = 0.007.
a2/0.5 = c2/0.3 = 0.15(0.5+0.3),
whence a2 = 0.5(0.15/0.8) = 0.094,
c2 = 0.3(0.15/0.8) = 0.056.
Thus A gains a1+a2 = 0.013+0.0.094 = 0.107 and C gains c1+c2 = 0.007+0.056 = 0.063.

$m_{12}(B \cap E)=0.02$ is redistributed to B and E using the PCR5:
b/0.2 = e/0.1 = 0.02/(0.2+0.1),
whence b = 0.2(0.02/0.3) = 0.013,
e = 0.1(0.02/0.3) = 0.007.
Thus B gains 0.013 and E gains 0.007.

Then one sums the masses of the conjunctive rule **m₁₂** and the redistribution of conflicting masses **m_r** (according to the information we have on each intersection, model, and relationship between conflicting hypotheses) in order to get the mass of the Unification of Fusion Theories rule $m_{UFT}$.

**m_UFT**, the Unification of Fusion Theories rule, is a combination of many rules and gives the optimal redistribution of the conflicting mass for each particular problem, following the given model and relationships between hypotheses; this extra-information allows the choice of the combination rule to be used for each intersection. The algorithm is presented above.
**m_lower**, the lower bound believe assignment, the most pessimistic/prudent belief, is obtained by transferring the whole conflicting mass to the total ignorance (Yager's rule) in a closed world, or to the empty set (Smets' TBM) in an open world herein meaning that other hypotheses might belong to the frame of discernment.

**m**<sub>middle</sub>, the middle believe assignment, half optimistic and half pessimistic, is obtained by transferring the partial conflicting masses $m_{12}(X \cap Y)$ to the partial ignorance $X \cup Y$ (as in Dubois-Prade theory or more general as in Dezert-Smarandache theory).

Another way to compute a middle believe assignment would be to average the **m**<sub>lower</sub> and **m**<sub>upper</sub>.

**m**<sub>upper</sub>, the lower bound believe assignment, the most optimistic (less prudent) belief, is obtained by transferring the masses of intersections (empty or non-empty) to the elements in the frame of discernment using the PCR5 rule of combination, i.e. $m_{12}(X \cap Y)$ is split to the elements X, Y (see Table 2). We use PCR5 because it is more exact mathematically (following the backwards the tracks of the conjunctive rule) than Dempster's rule, minC, and PCR1-4.

| X | $m_{12}(X)$ | A | B | C | D | E |
|---|---|---|---|---|---|---|
| A∩B | 0.040 | 0.020 | 0.020 | | | |
| A∩C | 0.170 | 0.107 | | 0.063 | | |
| A∩D | 0.200 | 0.111 | | | 0.089 | |
| A∩E | 0.090 | 0.020 | | | | 0.020 |
| | | 0.042 | | | | 0.008 |
| B∩C | 0.060 | | 0.024 | 0.036 | | |
| B∩D | 0.080 | | 0.027 | | 0.053 | |
| B∩E | 0.020 | | 0.013 | | | 0.007 |
| C∩D | 0.040 | | | 0.008 | 0.032 | |
| C∩E | 0.070 | | | 0.036 | | 0.024 |
| | | | | 0.005 | | 0.005 |
| D∩E | 0.080 | | | | 0.053 | 0.027 |
| Total | 0.850 | 0.300 | 0.084 | 0.148 | 0.227 | 0.091 |

Table 2. Redistribution of the intersection masses to the singletons A, B, C, D, E using the PCR5 rule only, needed to compute the upper bound belief assignment $m_{upper}$.

### 7.2. Negation/Complement Example:

Let $\Theta = \{A, B, C, D\}$ be the frame of discernment. Since $(\Theta, \cup, \cap, \mathcal{C})$ is Boolean algebra, the super-power set $S^\Theta$ includes complements/negations, intersections and unions. Let's note by $\mathcal{C}(B)$ the complement of B.

| | A | B | D | $\mathcal{C}(B)$ | A∩C | B∪C=B | A∩B | A∩$\mathcal{C}(B)$=A |
|---|---|---|---|---|---|---|---|---|
| | | | | | {Later in the dynamic fusion process we find out that A∩C is empty} | | | |
| | | | | | = φ | | = φ | ≠ φ |
| | | | | | Unknown relationship between A and C | | Optimistic in both A and B. | Consensus between A and $\mathcal{C}(B)$, but A⊂$\mathcal{C}(B)$ |
| $m_1$ | 0.2 | 0.3 | 0 | 0.1 | 0.1 | 0.3 | | |
| $m_2$ | 0.4 | 0.1 | 0 | 0.2 | 0.2 | 0.1 | | |
| $m_{12}$ | 0.08 | 0.09 | 0 | 0.02 | 0.17 | 0.03 | 0.14 | 0.08 |
| | | | | | ↓ | ↓ | ↓ | ↓ |
| | | | | | A∪C | B | A,B | A |
| $m_r$ | | | | | 0.17 | 0.03 | 0.082, 0.058 | 0.08 |
| **$m_{UFT}$** | **0.277** | **0.318** | **0.035** | **0.020** | **0** | **0** | **0** | **0** |
| $m_{lower}$ (closed world) | 0.16 | 0.26 | 0 | 0.02 | 0 | 0 | | |
| $m_{lower}$ (open world) | 0.16 | 0.26 | 0 | 0.02 | 0 | 0 | | |
| $m_{middle}$ (default) | 0.16 | 0.26 | 0 | 0.02 | 0 | 0 | | |
| $m_{upper}$ | 0.296 | 0.230 | 0 | 0.126 | 0.219 | 0.129 | | |

Table 3. Negation/Complement Example using the Unified Fusion Theories rule regarding a mixed redistribution of partial conflicting masses (Part 1).

| | A∩(B∪C) | B∩𝒞(B) | B∩(A∩C) | 𝒞(B)∩(A∩C) | 𝒞(B)∩(B∪C) = 𝒞(B)∩C | B∪(A∩C) =B |
|---|---|---|---|---|---|---|
| | = φ | = φ | = φ | = φ | = φ | |
| | At least one is right between A and B∪C | B is right, 𝒞(B) is wrong | No relationship known between B and A∩C (default case) | Very pessimistic on 𝒞(B) and A∩C | Neither 𝒞(B) nor B∪C are right | |
| $m_1$ | | | | | | |
| $m_2$ | | | | | | |
| $m_{12}$ | 0.14 | 0.07 | 0.07 | 0.04 | 0.07 | |
| | ↓ | ↓ | ↓ | ↓ | ↓ | ↓ |
| | A∪(B∪C) | B | B∪(A∩C)=B | A∪B∪C∪D | A, D | B |
| $m_r$ | 0.14 | 0.07 | 0.07 | 0.04 | 0.035, 0.035 | |
| **$m_{UFT}$** | **0** | **0** | **0** | **0** | **0** | **0** |
| $m_{lower}$ (closed world) | | | | | | |
| $m_{lower}$ (open world) | | | | | | |
| $m_{middle}$ (default) | | | | | | |
| $m_{upper}$ | | | | | | |

Table 3. Negation/Complement Example using the Unified Fusion Theories rule regarding a mixed redistribution of partial conflicting masses (Part 2).

|   | A∪B | A∪C | A∪D | B∪C | B∪D | C∪D | A∪B∪C | A∪B∪C∪D | φ |
|---|---|---|---|---|---|---|---|---|---|
|   |   |   |   |   |   |   |   |   |   |
|   |   |   |   |   |   |   |   |   |   |
| $m_1$ |   |   |   |   |   |   |   |   |   |
| $m_2$ |   |   |   |   |   |   |   |   |   |
| $m_{12}$ |   |   |   |   |   |   |   |   |   |
|   |   |   |   |   |   |   |   |   |   |
| $m_r$ |   |   |   |   |   |   |   |   |   |
| **$m_{UFT}$** | **0** | **0.170** | **0** | **0** | **0** | **0** | **0.140** | **0.040** | **0** |
| $m_{lower}$ (closed world) |   |   |   |   |   |   |   | 0.56 |   |
| $m_{lower}$ (open world) |   |   |   |   |   |   |   |   | 0.56 |
| $m_{middle}$ (default) | 0.14 | 0.17 |   | 0.03 |   |   | 0.14 | 0.11 |   |
| $m_{upper}$ |   |   |   |   |   |   |   |   |   |

Table 3. Negation/Complement Example using the Unified Fusion Theories rule regarding a mixed redistribution of partial conflicting masses (Part 3).

**Model of Negation/Complement Example:**

$A \cap B = \phi$, $C \subset B$, $A \subset \mathcal{C}(B)$.

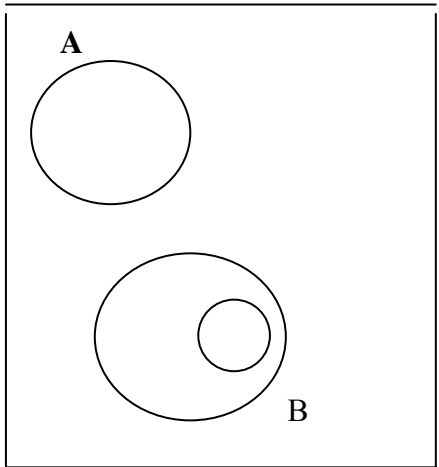

Fig. 1

$m_{12}(A \cap B) = 0.14$.
$x1/0.2 = y1/0.1 = 0.02/0.3$, whence $x1 = 0.2(0.02/0.3) = 0.013$, $y1 = 0.1(0.02/0.3) = 0.007$;
$x2/0.4 = y2/0.3 = 0.12/0.7$, whence $x2 = 0.4(0.12/0.7) = 0.069$, $y2 = 0.3(0.12/0.7) = 0.051$.
Thus, A gains $0.013+0.069 = 0.082$ and B gains $0.007+0.051 = 0.058$.

For the upper belief assignment $m_{upper}$ one considered all resulted intersections from results of the conjunctive rule as empty and one transferred the partial conflicting masses to the elements involved in the conflict using PCR5.
All elements in the frame of discernment were considered non-empty.

### 7.3. Example with Intersection:

Look at this:

Suppose A={x<0.4}, B={0.3<x<0.6}, C={x>0.8}. The frame of discernment T={A, B, C} represents the possible cross section of a target, and there are two sensors giving the following bbas:
m1(A)=0.5, m1(B)=0.2, m1(C)=0.3.
m2(A)=0.4, m2(B)=0.4, m2(C)=0.2.

|            | A   | B   | C   | A∩B= {.3<x<.4} | A∪C | B∪C |
|------------|-----|-----|-----|----------------|-----|-----|
| $m_1$      | .5  | .2  | .3  |                |     |     |
| $m_2$      | .4  | .4  | .2  |                |     |     |
| $m_1$&$m_2$ DSmT | .20 | .08 | .06 | .28            | .22 | .16 |

We have a DSm hybrid model (one intersection A&B=nonempty ).
This example proves the necessity of allowing intersections of elements in the frame of discernment.   [Shafer's model doesn't apply here.]
Dezert-Smarandache Theory of Uncertain and Paradoxist Reasoning (DSmT) is the only theory which accepts intersections of elements.

### 7.4. Another Multi-Example of UFT:

Cases:
1. Both sources reliable: use conjunctive rule [default case]:
  1.1. A∩B≠ϕ:
    1.1.1. Consensus between A and B; mass → A∩B;
    1.1.2. Neither A∩B nor A∪B interest us; mass → A, B;
  1.2. A∩B=ϕ:
        1.2.1. Contradiction between A and B, but optimistic in both of them; mass → A, B;
        1.2.2. One right, one wrong, but don't know which one; mass → AχB;
    1.2.3. Unknown any relation between A and B [default case]; mass → AχB;
    1.2.4. Pessimistic in both A and B; mass → AχB;
    1.2.5. Very pessimistic in both A and B;
    1.2.5.1. Total ignorance ε AχB; mass → AχBχCχD (total ignorance);
    1.2.5.2. Total ignorance = AχB; mass → ϕ (open world);
    1.2.6. A is right, B is wrong; mass → A;

1.2.7. Both A and B are wrong; mass → C, D;
1.3. Don't know if A∩B = or ≠ φ (don't know the exact model); mass → A∩B (keep the mass on intersection till we find out more info) [default case];
2. One source reliable, other not, but not known which one: use disjunctive rule; no normalization needed.
3. S1 reliable, S2 not reliable 20%: discount S2 for 20% and use conjunctive rule.

|  | A | B | A∪B | A∩B | φ (open world) | A∪B∪C∪D | C | D |
|---|---|---|---|---|---|---|---|---|
| S1 | .2 | .5 | .3 | | | | | |
| S2 | .4 | .4 | .2 | | | | | |
| S1&S2 | .24 | .42 | .06 | .28 | | | | |
| S1 or S2 | .08 | .20 | .72 | 0 | | | | |
| UFT 1.1.1 | .24 | .42 | .06 | .28 | | | | |
| UFT 1.1.2 (PCR5) | .356 | .584 | .060 | 0 | | | | |
| UFT 1.2.1 | .356 | .584 | .060 | 0 | | | | |
| UFT 1.2.2 | .24 | .42 | .34 | 0 | | | | |
| UFT 1.2.3 | .24 | .42 | .34 | 0 | | | | |
| UFT 1.2.4 | .24 | .42 | .34 | 0 | | | | |
| UFT 1.2.5.1 | .24 | .42 | .06 | 0 | 0 | .28 | | |
| UFT 1.2.5.2 | .24 | .42 | .06 | 0 | .28 | | | |
| 80% S2 | .32 | .32 | .16 | | | .20 | | |
| UFT 1.2.6 | .52 | .42 | .06 | | | | | |
| UFT 1.2.7 | .24 | .42 | .06 | 0 | | | .14 | .14 |
| UFT 1.3 | .24 | .42 | .06 | .28 | | | | |
| UFT 2 | .08 | .20 | .72 | 0 | | | | |
| UFT 3 | .232 | .436 | .108 | .224 | | 0 | | |

**8. Acknowledgement**.
We want to thank Dr. Wu Li from NASA Langley Research Center, Dr. Philippe Smets from the Université Libre de Bruxelles, Dr. Jean Dezert from ONERA in Paris, and Dr. Albena Tchamova from the Bulgarian Academy of Sciences for their comments.